\documentclass[review]{elsarticle}

\journal{Journal of \LaTeX\ Templates}

\usepackage{mathtools}
\usepackage{amsthm}
\usepackage{subcaption}
\usepackage{paralist}
\usepackage{bm}
\usepackage{amsfonts}
\usepackage{multirow}
\usepackage{cases}
\usepackage{booktabs}
\usepackage{threeparttable}
\usepackage{epstopdf}
\usepackage{upgreek}
\usepackage{endnotes}
\usepackage{etoolbox}
\usepackage{algpseudocode}
\usepackage{amsmath}
\usepackage{color}
\usepackage{algorithm}
\usepackage{bbding}
\usepackage[colorlinks,citecolor=blue]{hyperref}

\begin{document}
	
	\begin{frontmatter}
		
		\title{DCANet: Dense Context-Aware Network for Semantic Segmentation}

		\author[mymainaddress]{Yifu Liu \corref{mycorrespondingauthor}}
		\ead{liuyf@zju.edu.cn}
		\author[mymainaddress]{Chenfeng Xu}
		%\author[mymainaddress]{Chao Chen}
		%\author[mymainaddress]{Zhihong Chen}
		\author[mymainaddress]{Xinyu Jin}
		\cortext[mycorrespondingauthor]{Corresponding author}
		
		\address[mymainaddress]{Institute of Information Science and Electronic Engineering, Zhejiang University, Hangzhou, China}
		
		\begin{abstract}
			As the superiority of context information gradually manifests in advanced semantic segmentation, learning to capture the compact context relationship can help to understand the complex scenes.
			In contrast to some previous works utilizing the multi-scale context fusion, we propose a novel module, named \textit{Dense Context-Aware} (DCA) module, to adaptively 
			integrate local detail information with global dependencies. %through a more efficient way
			Driven by the contextual relationship, the DCA module can better achieve the aggregation of context information to generate more powerful features. %in different 
			Furthermore, we deliberately design two extended structures based on the DCA modules to further capture the long-range contextual dependency information. By combining the DCA modules in cascade or parallel, our networks use a progressive strategy to improve multi-scale feature representations for robust segmentation.
			We empirically demonstrate the promising performance of our approach (DCANet) with extensive experiments on three challenging datasets, including PASCAL VOC 2012, Cityscapes, and ADE20K.
		\end{abstract}
		
		\begin{keyword}
			Semantic segmentation \sep Dense Context-Aware module \sep Long-range contextual information \sep Progressive strategy
		\end{keyword}
		
	\end{frontmatter}
	
%	\linenumbers
	
	\section{Introduction}
	\label{intro}
	Semantic segmentation is a fundamental visual understanding task by solving a dense labeling problem, whose purpose is to assign 
	different semantic categories for each pixel of the given image. 
	%重要的价值被展现出来 successfullyc  \cite{8700608}\cite{Zhang2019}   \cite{9165167} \cite{7293666}
	It plays a crucial role in many potential applications such as scene understanding, autonomous driving, and human parsing, etc.
	%has been actively employed in many potential applications such as scene understanding, autonomous driving, and human parsing, etc. %to name a few.
	%etc., but still has some tricky issues. %valuable
	%It has currently shown many potential application values such as scene understanding, human parsing, and autonomous driving, etc, but still has some tricky problems.  %challenges
	
	%这一段先说一下state-of-art, 然后再说其中很重要的一点就是探索context,  integrate   to get long-range dependencies as well as local boundary information.
	Recently, a lot of state-of-the-art semantic segmentation techniques based on the Fully Convolutional Network (FCN) \cite{Shelhamer2017} have achieved striking results.
	% on the semantic segmentation task. %frameworks
	Among these works, one of the most effective approaches to enhance the robustness of segmentation is exploiting the scene context information to boost understanding of the various objects in semantic level.
	For example, some works \cite{Peng_2017_CVPR}\cite{Zhang_2018_CVPR} enlarge the kernel size with a decomposed structure or propose an effective context encoding layer to obtain rich global contextual information. And some works \cite{Liu15parsenet}\cite{Chen_2017_CVPR}\cite{Zhao_2017_CVPR}\cite{ZHAO2019273} aggregate multi-scale contextual information in the form of different pooling and dilated convolution operations to capture the correlation between regions. %get calculate context.
	%The proposed pyramid pooling module in PSPNet and atrous spatial pyramid pooling (ASPP) module in DeepLab not only help extract the context information to a certain degree, but also provide a parallel method of the aggregation of multi-scale features.
	The pyramid pooling module in PSPNet \cite{Zhao_2017_CVPR} and atrous spatial pyramid pooling (ASPP) module in DeepLab \cite{Chen_2015_ICLR}\cite{Yu_2016_ICLR}\cite{Chen_2017_CVPR} also provide effective parallel processing manners to help extract multi-scale context features to a certain degree.
	Besides, to establish long-range dependencies, the encoder-decoder structures \cite{UNET_2015}\cite{Lin_2017_CVPR}\cite{Ding_2018_CVPR} combine low-level and high-level semantic features by applying skip connections \cite{He_2016_CVPR}, which also achieves the multi-level context fusion. %[] to achieve the multi-level context fusion.

\begin{figure}[!ht]
		% Use the reqleqqant command to insert your figure file.
		\centering
		\includegraphics[width=1.0\linewidth]{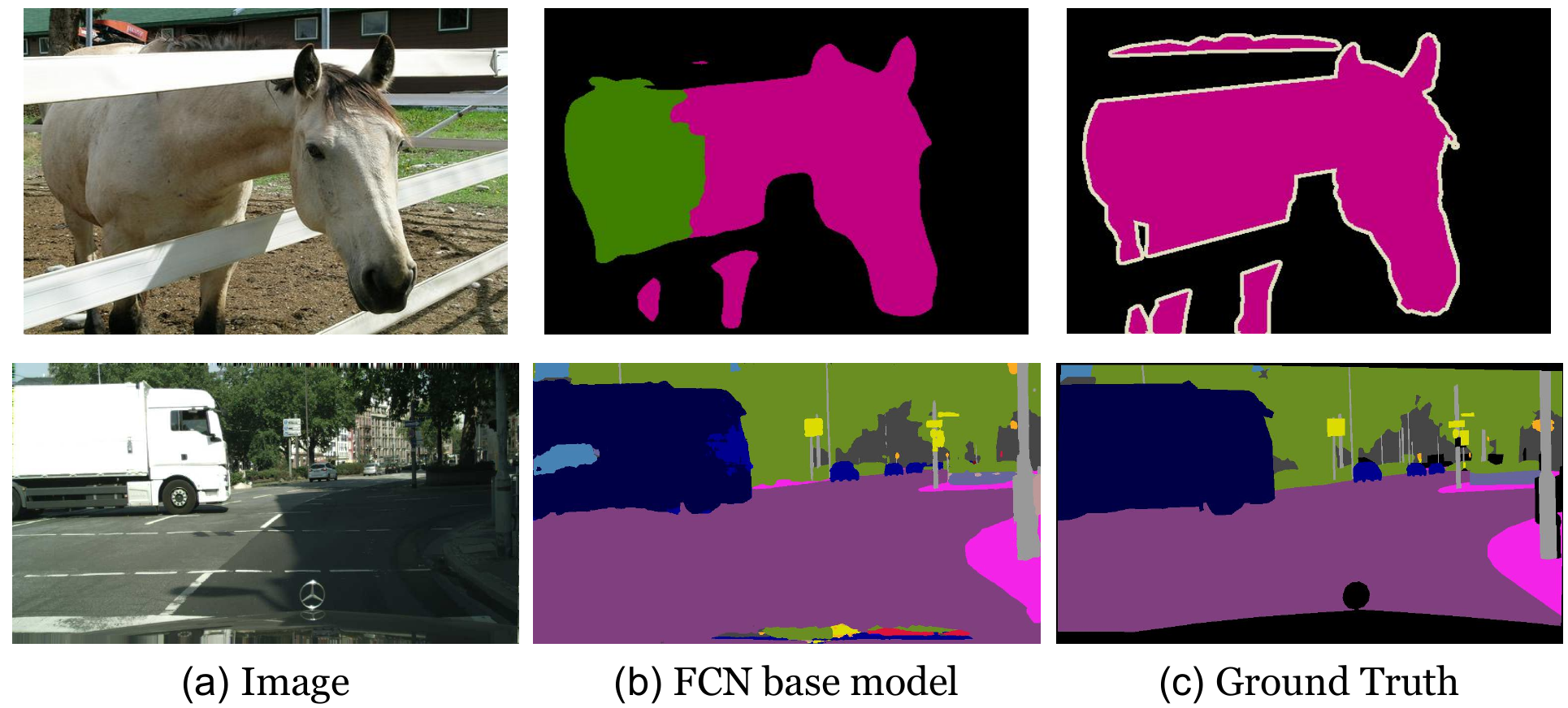}
		\caption{Illustration of hard examples from different datasets. (a) Input image. (b) the output of FCN based model. (c) GroundTruth label.
			These examples are from the PASCAL VOC 2012 and the Cityscapes dataset. In the first row, the left corner of the horse is recognized as a cow.
			In the second row, there are some obvious misclassified categories on the truck.
			The various scales, occlusion and similar appearance of objects need to dense contextual information to parse each pixel.
			%导致了上述问题！    这个cityscapes的图可能要换！
		}
		\label{Fig1}            
	\end{figure}

	%But it lacks a portrayal of semantic information and detailed information. But it lacks a description of the feature
	%Although the context fusion used in these approaches can learn to generalize the different scales objects in the image, it can not capture the deep relationship of different context beacuse of just adopting the concatenation of different features. 
	Although the context fusion used in these approaches learns to generalize different scales objects to some extent, it can not leverage the relationship between objects in a global view, which is essential to capture the long-range contextual relationship for scene understanding \cite{Yuan_18_ocnet}\cite{PSANet_2018}. 
	Consequently, the segmentation networks with theses limited fusion methods such as concatenation or addition operation between global and local clues, have no sufficient understanding of objects in the scene. %resulting in misclassification especially for some similar categories. 
	As a result, misclassification could occur, especially for some similar categories.
	%Due to the limitations of aggregating context information, such as just concatenation or addition and treating all pixels equally, the segmentation networks have an insufficient understanding of objects in the scene and generate the misclassified results, especially for some similar categories.
	%some indistinguishable samples. such as just adopting the concatenation of different features,
	To better illustrate the aforementioned issues, we show several representative examples in Fig. \ref{Fig1}.
	%To better illustrate the situations above, we show a few representative examples in Fig. \ref{Fig1}. %{\color{red}which contains the segmentation results from our experiments.} %make some comparison in Fig. \ref{Fig1}, 
	Intuitively, we need to adaptively establish features dependencies in the spatial and channel dimensions to enhance the awareness of the scene context and collect the long-range dependencies from all pixels. % capture
	%或者这句话改成说明他们缺失这种依赖导致问题,  或者这里建立改成捕捉或自适应的寻找   progressively
	
	Towards the above issues, we design a Dense Context-Aware (DCA) module, a novel short-range module that captures the compact context relationship and explicitly enhances the capabilities of networks for multi-scale processing. %first propose the
	%the processing of multi-scale features. 
	Instead of directly utilizing the multi-scale context aggregation \cite{Liu15parsenet}\cite{Zhao_2017_CVPR}\cite{Lin_2017_CVPR}, our proposed DCA module extracts global descriptors and local features from two pathways under the contextual correlation.
	%Instead of aggregating multi-scale features in a simple manner, such as concatenation and addition, and treating all pixels equally, our dense context-aware modules extract global {\color{red}view} and local features from two pathways under the contextual correlation.
	%utilizes two pathways with local and more global context to enhance contextual awareness of semantic features. %这里参考autofocus layer
	%we use the Dense Context-Aware module to enhance contextual awareness of semantic features and {\color{red}complete the fusion of different features in corresponding forms}.
	%be aware of different context regions. Enhance contextual awareness of semantic features
	In detail, we regard the semantic level features as dense contextual awareness to adaptively apply attention to local clue in a data-driven manner, after which we apply the concatenation and addition operation to complete the fusion of different features, respectively.
	%complete the fusion of different features in corresponding forms.
	%merge the context features to generate two different branches.
	
	%By combining these modules, our network can progressively refine richer context information for a comprehensive understanding of complex scenes.
	%To 为了建立这种长期地关系我们需要利用多个模块来维持, Concretely,
	%受到psp ...的启发，我们还进行了some extensions   Moreover ,
	Furthermore, we carry out a sequence of DCA modules at multiple scales to capture long-range contextual dependencies. %Concretely, 
	Based on the DCA module, we further put forward two extended structures: (i) a novel network structure named Cascade-DCA, which consists of several DCA modules working at different scales context. %And we put these modules into a cascaded network in an appropriate way.
	And we connect these modules in a cascading manner to get a finer long-range relationship.
	%use cascading operations to connect these modules to improve the features.
	(ii) the Pyramid-DCA structure, following the pyramid design introduced in PSPNet, which performs multiple branches, each one utilizes cascaded DCA modules and finally accomplish feature fusion from different branches.
	%fuses these features 
	With the long-range dependencies established, our network can progressively refine spatial relationships from a global view and improve feature representations for a comprehensive perception of complex scenes. 
	%With establishing the long-range dependencies, our network can progressively refine spatial relationships from a global view and improve feature representations for a comprehensive perception of complex scenes. %dense contextual richer context information
	
	We extensively evaluate our DCANet on three most competitive semantic segmentation datasets, \textit{i.e.}, PASCAL VOC 2012, Cityscapes, and ADE20K.
	Experimental results show that our proposed method consistently outperforms strong baselines and obtains significant results.
	%We give all implementation details related to our decent performance in this paper, and make the code and trained models publicly available to the community\footnote{https://github.com/YifuLiuL/DCANet}. 
	We will give all implementation details related to our decent performance in this paper and will make the code and trained models publicly available to the community upon publication of the paper with a license that allows free usage for research purposes.%\footnote{https://github.com/YifuLiuL/DCANet}
	Our main contributions can be summarized as follows:
	\begin{itemize}
		%第一条可以说一下注意力 这里还得斟酌一下语言 better fuse semantic information and localization information for  from the intermediate features.
		\item We first improve the manner of the context fusion and propose a novel network module named Dense Context-Aware (DCA), taking advantages of the contextual awareness to aggregate the features from the perspective of position and channel. %and highlight their feature representations.   which can
		%And our module can be plugged into any fully convolutional neural network for exploiting different network variants
		%	generate new features with dense and rich contextual information. {\color{red}Based on the DCA module, our flexible design allows for a simple exploration of different network variants. }    %这句话可能要放到别的地方
		%第二条想下要不要说对比各种形式
		\item We further propose extended structures based on our DCA modules to model dense contextual guidance at several scales in a more efficient way.
		%The segmentation results of our DCANet are improved with capturing the semantic similarity and long-range contextual dependency.
		By capturing the semantic similarity and long-range contextual dependency, our DCANet improves the segmentation results.
		%achieves a better comprehensive understanding of complex scenes.
		%get discriminative and effective features by fusing differnent scale context features selectively. %increase discriminating and presentation ability
		%	\item A dual-stream structure with different concerns is proposed for better realizing the decomposition of tasks. And our scale context selection attention modules further improve the segmentation results by modeling rich contextual dependencies of global features and local features.
		
		%最后一条写泛化和效果，以及优势可以嵌入
		\item Extensive experiments on three challenging semantic segmentation datasets, including PASCAL VOC 2012, Cityscapes and ADE20K, demonstrate the superiority of our approach over other previous state-of-the-art methods. More importantly, we visualize the attention maps of our DCA module to enhance the interpretability of deep CNNs. %demonstrate the {\color{red}satisfying performance} of our approach. 
		%	Our experimental results show a satisfying performance on segmentation tasks. Furthermore, we visualized the attention maps of deep CNNs to enhance the interpretability of scale context selection attention module, and this unveils the black box network operation for the role of features to some extent.
	\end{itemize}
	The rest of the paper is organized as follows. We first discuss related work in Section \ref{sec:2}. 
	After introducing our approach and network architectures in detail in Section \ref{approach}, we present experimental results in Section \ref{Exper}. 
	Finally, we briefly conclude the paper in Section \ref{conc}.
	
	%\paragraph{Installation} If the document class \emph{elsarticle} is not available on your computer, you can download and install the system package \emph{texlive-publishers} (Linux) or install the \LaTeX\ package \emph{elsarticle} using the package manager of your \TeX\ installation, which is typically \TeX\ Live or Mik\TeX.
	%
	%\paragraph{Usage} Once the package is properly installed, you can use the document class \emph{elsarticle} to create a manuscript. Please make sure that your manuscript follows the guidelines in the Guide for Authors of the relevant journal. It is not necessary to typeset your manuscript in exactly the same way as an article, unless you are submitting to a camera-ready copy (CRC) journal.
	%
	%\paragraph{Functionality} The Elsevier article class is based on the standard article class and supports almost all of the functionality of that class. In addition, it features commands and options to format the
	%\begin{itemize}
	%\item document style
	%\item baselineskip
	%\item front matter
	%\item keywords and MSC codes
	%\item theorems, definitions and proofs
	%\item lables of enumerations
	%\item citation style and labeling.
	%\end{itemize}
	\section{Related Work}
	\label{sec:2}
	In the following, we review recent advances related to our method. For other ways, a comprehensive review can be found in \cite{LATEEF2019321}.\\

	\noindent\textbf{Semantic Segmentation. } %\cite{FU2020107152}\cite{article2012}\cite{2014Very}\cite{He_2016_CVPR}
	Recently, benefiting from the advances of deep neural networks, semantic segmentation or scene parsing has achieved significant progress. Fully Convolutional Network (FCN) was the first approach to use convolution layers to replace the fully-connected layer for semantic segmentation. DeconvNet \cite{Noh2015}, SegNet \cite{Badrinarayanan2017} and RefineNet\cite{Lin_2017_CVPR}, etc., adopted the encoder-decoder structure to combine low-level and high-level information for the optimization of the segmentation results.
	Besides, Markov random field (MRF) \cite{Liu_2015_ICCV} and conditional random field (CRF) \cite{Chen_2015_ICLR}\cite{chen_2018_deeplabv2}\cite{Zheng_2015_ICCV} are broadly utilized to model the long-range dependencies for fine structure prediction in semantic segmentation.
	%Besides, DeepLab \cite{chen_2018_deeplabv2} and CRF-RNN \cite{Zheng_2015_ICCV} utilized the Conditional random field (CRF) to model the long-range dependencies for fine structure prediction in semantic segmentation.           the works in
	To obtain a larger receptive field of convolutional neural networks, Chen \textit{et al.} \cite{Chen_2015_ICLR}\cite{chen_2018_deeplabv2} employed the dilated convolution operation to enlarge the spacing of values (insert 'holes') while increasing the feature resolution. In our work, we also employ the same dilated strategy as in \cite{Chen_2017_CVPR}\cite{Zhang_2019_ICCV} to preserve the intermediate features of high-resolution. \\
	
	\noindent\textbf{Context. }%\cite{ZHOU2016312}
	The context always plays an important role in various computer vision tasks. 
	Contextual information, with various forms such as global scene context and sampled spatial context, has been applied for image classification \cite{GonzalezGarcia_2018_CVPR} and object detection \cite{Liu_2018_CVPR}, especially in the semantic segmentation.
	%With various forms such as global scene context and sampled spatial context, the contextual information has been applied for image classification \cite{Gonzalez-Garcia_2018_CVPR}\cite{Thacker1990} and object detection \cite{Liu_2018_CVPR}, especially in the semantic segmentation.
	Works like \cite{Szegedy_2015_CVPR}\cite{2020Deep}\cite{He_2016_CVPR}\cite{LIAN2020454} used the global average pooling (GAP) to obtain the global contextual prior.
	%The works in \cite{2014Very}\cite{Szegedy_2015_CVPR}\cite{He_2016_CVPR} used the global average pooling (GAP) to obtain the global contextual prior. 
	Moreover, Liu \textit{et al.} introduced ParseNet \cite{Liu15parsenet} that applies global pooling to attain the fusion of context information for scene parsing, and Zhao \textit{et al.} proposed PPM \cite{Zhao_2017_CVPR} module to fuse multi-scale contextual information.
	The atrous spatial pyramid pooling (ASPP) \cite{chen_2018_deeplabv2} was developed to aggregate contextual information by different dilated rates.
	%考虑一下要不要介绍ACFNet 的class context
	In addition, Zhang \textit{et al.} \cite{Zhang_2019_ICCV} proposed a novel framework named Attentional Class Feature Network (ACFNet), to harvest the contextual information from a categorical perspective.
	
	%这里可以具体到啥方法 \cite
	In reference to the above method but different from them, we aim to use the proposed dense context-aware (DCA) module to progressively achieve the transition from short-range attention to long-term contextual dependency. \\
	%mainly aims to progressively harvest the long-range contextual dependency aggregate the context information more effectively. 

	\noindent\textbf{Attention. }
	Attention mechanism is widely used in deep neural networks and has achieved excellent performance.
	%别忘了加一个自己地模块 Liu \textit{et al.} 
	Mnih \textit{et al.} \cite{NIPS2014_09c6c378} introduced an attention model that adaptively selects a sequence of regions or locations and only processes the selected regions.
	Chen \textit{et al.} \cite{Chen_2016_CVPR} learned several attention masks from different network branches to fuse weighted feature maps.
	Squeeze-and-Excitation Network (SENet) \cite{Hu_2018_CVPR} was used to improve the representational power of deep features by modelling channel-wise relationships in an attention mechanism.
	Wang \textit{et al.} \cite{Wang_2018_CVPR} brought forward the non-local module for vision tasks to calculate the spatial-temporal dependencies through the self-attention form.
	OCNet \cite{Yuan_18_ocnet} and DANet \cite{Fu_2019_CVPR} utilized the self-attention mechanism to harvest the related contextual information.
	PSANet \cite{PSANet_2018} also learned an attention map to aggregate contextual information for each individual point adaptively and specifically.

	Our attention module is motivated by the success of attention mechanisms in the above works. We rethink the attention mechanism from the perspective of the context fusion and compute the attention masks about the global information for better improving the feature representations. %the capability of features.
	
	\section{Approach}
	\label{approach}
	
	In this section, we first present a general framework of our network and introduce the key component---dense context-aware (DCA) module with the details of the specific formulations and operations.
	Then we elaborate on two long-range dense context-aware structures that we propose based on the DCA module.
	%our proposed xxx (DCA) module with the details of the specific formulations and operations and elaborate several variant networks that we propose based on the DCA module.
	Finally, we describe the complete network structures of our method.
	%structures, which consist of our DCA modules.} %named DCANet.
	
	\subsection{Overview}
\begin{figure}[!ht]
		\centering
		\includegraphics[width=1.0\linewidth]{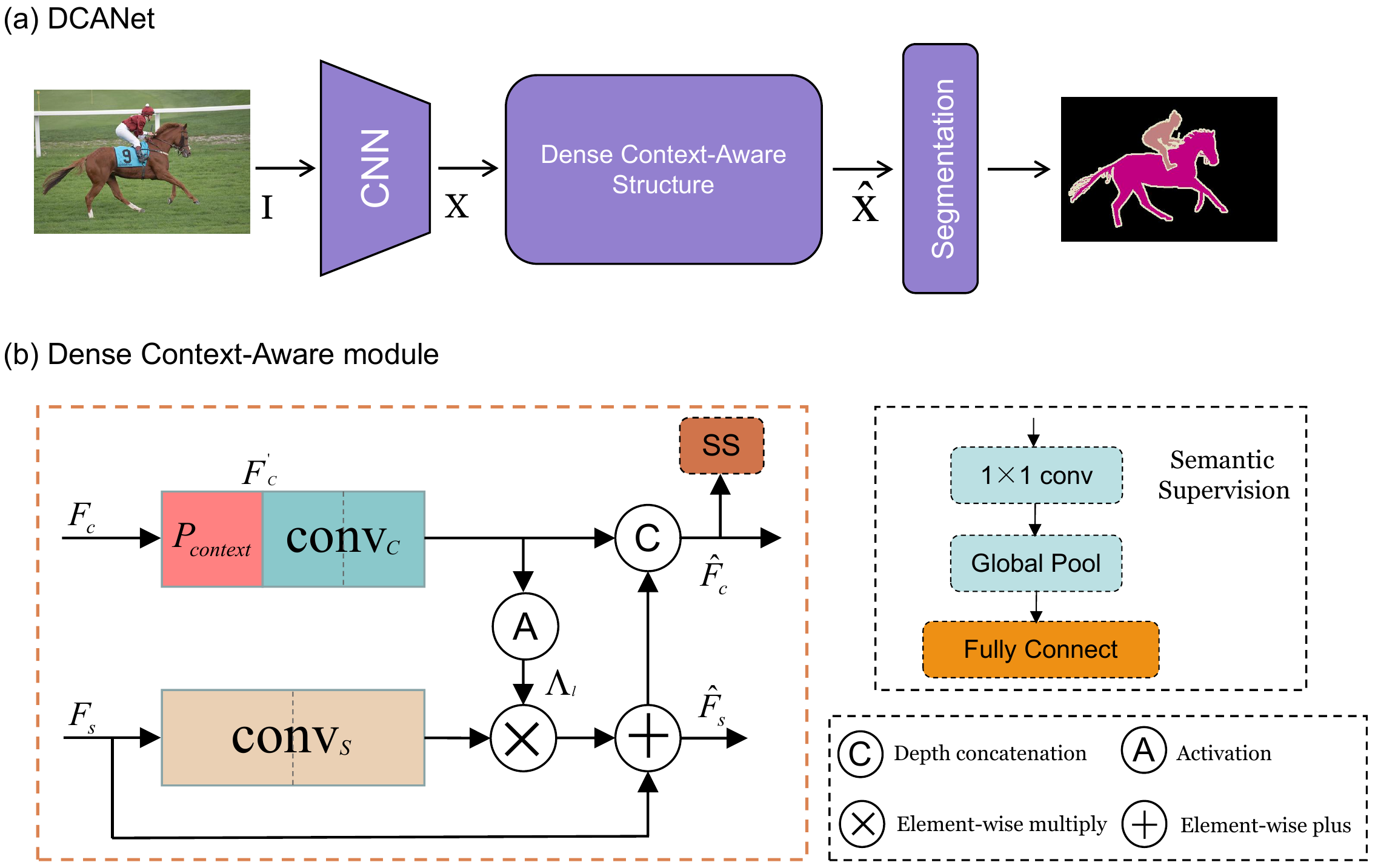}
		\caption{An overview of the proposed DCANet for semantic segmentation. (a) The overall network architecture: Given an input image, we first use a fully convolution network (FCN) to get the feature map $\rm X \in \mathbb{R}^{C \times H \times W}$, then employ the long-range dense context-aware (DCA) structure and output an updated feature map $\rm \hat{X} \in \mathbb{R}^{C^{\prime} \times H \times W}$. Based on the updated feature map, we employ the segmentation layer, which predicts the pixel-wise score map and upsamples the score map for 8x times by the bilinear interpolation, to generate the final segmentation map.
			(b) The details of Dense Context-Aware module. \textbf{SS} --- semantic supervision. } %where different scale features are {\color{red}carried} in each stream.}
		\label{Fig2}       % Give a unique labelwe employ the segmentation layer to generate the final segmentation map.
	\end{figure}

	An input image ${\rm I}$ is fed into a fully convolution network (\textit{e.g.}, a ResNet backbone) to acquire the feature map ${\rm X}$ . % of size $W \times H$.
	%Given an input imgae ${\rm I}$, our proposed approach feeds the image ${\rm I}$ to a fully convolution network (\textit{e.g.}, a ResNet backbone) to acquire the feature
	%map ${\rm \hat{X}}$ of size $W \times H$, 
	And then our approach lets the feature map ${\rm X}$ go through the long-range dense context-aware structure, generating a processed feature map ${\rm \hat{X}}$.
	Finally, the segmentation layer predicts the category of each pixel based on the generated feature map ${\rm \hat{X}}$, and upsamples the score map for eight times at last. 
	The pipeline is presented in Fig. \ref{Fig2}(a), and the whole structure is called DCANet. 
	The critical contribution of DCANet to semantic segmentation lies in the long-range DCA structure, which is mainly composed of dense context-aware modules. 
	%{\color{red}Finally, the fused features are fed into the segmentation layer to generate the final segmentation map.}
	% 要不要这里我们一直去说明他们属于哪个空间 C*H*W
	%$\{\mathbf{Q}, \mathbf{K}\} \in \mathbb{R}^{C^{\prime} \times W \times H}$
	
	\subsection{Dense Context-Aware Module}
	\label{sec:3.2}%better achieve the fusion of contextual information by capturing compact context relationship at the pixel level.
	The intuition of dense context-aware module is to enhance the contextual awareness of features by applying the related context attention to the local information.
	%To further improve the feature representation with the context fusion, the module employs appropriate fusion manners over global and local features. 
	To further improve the feature representation with the context fusion, our solution achieves through two aspects: fusion with the context attention and semantic supervision for later input features. \\
	% exploit 这个词也蛮不错的
	%In PSPNet [47], the average pooling is employed over four different pyramid scales and pixels in one sub-region are treated as the context of pixels within the same sub-region. Some other works focus on how to fuse different context information [43, 42, 12, 28] more selectively. In contrast to conventional context described above, in this paper, we harvest the contextual information from a categorical perspective.
	
	%说明目前的大多数方法 The various network architectures proposed to effectively capture image context can be broadly grouped into three categories.
	
	%The various methods proposed to complete the context fusion can be summarized as the fusion of different scale context.
	%Therefore, our DCA module also utilizes the {\color{red}multi-scale context} to further apply contextual attention at the pixel level.
	\noindent\textbf{Fusion with Context Attention. }
	Initially, in order to combine the local and global information for modelling rich contextual relationships, the DCA module comprises two pathways, the contextual pathway and the spatial pathway. % the combinations of local and global information to model rich contextual relationships, 
	%We use $F_{c}^{}$ and $F_{s}^{}$ to represent the two inputs of the DCA module, which are either the intermediate features or the two outputs of the previous module.
	%Note that we use the superscript letter to denote the pathway.
	We use $F_{c}^{}$ and $F_{s}^{}$ to represent the two inputs of the DCA module, where the subscript letters denote the pathway. In particular, we take the output of the CNN network as the input to each pathway of the first DCA module in the long-range DCA structure.
	%{\color{red}The inputs of the DCA module are $F_{c}^{}$ and $F_{s}^{}$, which are either the same intermediate features or the outputs of the previous module.}
	%corresponding to the contextual pathway and the spatial pathway, respectively.
	As the module shown in Fig. \ref{Fig2}(b), its detailed process is described by two steps.
	The first step is to calculate the attention masks over the contextual pathway.
	
	Inspired by \cite{Peng_2017_CVPR}\cite{Zhao_2017_CVPR}\cite{qin2018autofocus}, which were proposed to enlarge the receptive field on the input feature, we adopt the average pooling \cite{1641019} for generating the context priors in a global view.
	%Referring to various methods \cite{Peng_2017_CVPR}\cite{Zhao_2017_CVPR}\cite{autofocus_2018} proposed to enlarge the receptive field on the intermediate feature, we adopt the average pooling \cite{1641019} for generating the context priors in a global view.
	In the DCA module, this operation is named context pooling, of which the outputs' size is represented by the $r_{l}$.
	%whose pixel size of pooling can be represented by the $r_{l}$.
	%it can not leverage the relationship between objects or stuffs in a global view
	The input feature $F_{c}^{}$ in the context pathway first passes through the context pooling to generate the features $F_{c}^{'}$ with the specific scale context.
	%We can set the size of r for different modules, which is an important basis for achieving our long-term structure
	%By decreasing the size $r_{l}$ of scale context, the greater context but in less detail can be captured in the contextual pathway.
	The greater context but less detailed can be captured in the contextual pathway by decreasing the size $r_{l}$ of scale context.
	Thus, we can enlarge the receptive field via controlling $r_{l}$ that leads to a 'zoom out' behaviour over the features.
	The context pooling operation can be formulated as follows:
	%{\color{red}Thus the operation leads to a 'zoom out' behaviour, and the context pooling operation can be formulated as follows: % by adaptively capture greater context
	\begin{flalign}
	F_{c}^{'}=P_{context}\left(F_{c}^{} \ ; \ r_{l}\right) \label{Eq1}
	\end{flalign}	
	Here we can manually set the scale context size $r_{l}$ for different modules on the occasion of multiple modules used in the DCA structures (see Section \ref{sec:3.3}).
	%using multiple modules  following
	
	Meanwhile, we keep the original scale size for the features $F_{s}^{}$ which are rich in spatial details in the other pathway.
	%As shown in Fig. 3, specifically, in the DCA module, we apply a scale transformation operation for the intermediate feature map $X$ obtain the global context features $F_{0}^{C}$, and we keep the original scale size for the features $F_{0}^{S}$ of other pathway which are rich in spatial details.  % maintain
	%Note that the subscript number denotes the t-th attention modules, and the superscript letter represents the pathway.
	%with a normalization layer [11, 37] and ReLU [26] in between
	Then the context features $F_{c}^{'}$ goes through two convolutional layers with batch normalization and ReLU (the $\operatorname{conv}_{c}$ in Fig. \ref{Fig2}(b)), after which the features are further mapped by an element-wise sigmoid function to obtain the dense context attention masks $\Lambda_{l}$. To align with the other pathway, we apply an interpolation operation in the second convolutional layer to obtain the features of identical size as the original feature map. The process can be described in mathematical as follows: %Mathematically,
	\begin{flalign}
	\Lambda_{l}=\sigma\left(\operatorname{conv}_{c}\left(F_{c}^{'}\right)\right) \label{Eq2}
	\end{flalign} 
	% 突然想到可以分为两个step 1. calculation 2. Update
	The module also generates two outputs, which serve as the input of the next attention module under connecting multiple DCA modules. We explain the second step, updating the two outputs of the module.
	%{\color{red}Next, we explain the step of updating the two outputs of the module.}
	
	Having computed the context attention masks $\Lambda_{l}$, the spatial pathway $\hat{F}_{s}^{}$ is updated by:
	\begin{flalign}
	\hat{F}_{s}^{}=\Lambda_{l} \otimes \operatorname{conv}_{s}\left(F_{s}^{}\right)+F_{s}^{} \label{Eq3}
	\end{flalign}
	where $\otimes$ denotes an element-wise multiplication. By multiplying the spatial features after two convolution layers (the $\operatorname{conv}_{s}$ in Fig. \ref{Fig2}(b)) with the attention masks $\Lambda_{t}$, 
	we extract the dense pixel-wise prediction maps that are driven by the contextual awareness for adaptively achieving the optimal representation of features.
	%can adaptively improve the contextual awareness for achieving the aggregation of multiple features in a data-driven manner.
	Considering the preservation of the original spatial information, we employ the short-range residual connection to perform an element-wise plus. And the residual connection \cite{He_2016_CVPR} helps capture the long-range contextual relationship and achieves effective end-to-end training of the whole network, particularly when there are several DCA modules.
	
	As the spatial pathway gets updated through the DCA module, the contextual pathway should keep being updated with the other pathway, which can complete the transmission of spatial features to the following module. % By doing so, the updated step can accomplish the transmission of spatial features to the next module.
	Therefore, we incorporate the calculated spatial features with rich context information.
	The context features $F_{c}^{'}$ further pass through the two convolutional layers in the previous step.
	%{\color{red}Notably, we have mentioned in the previous step that the context features $F_{c}^{'}$ further pass through the two convolutional layers.   
	Then, the transformed features are combined with the updated spatial features $\hat{F}_{s}^{}$ by concatenation. Mathematically, the update can be formulated as follows:
	\begin{flalign}
	\hat{F}_{c}^{}=\operatorname{conv}_{c}\left(F_{c}^{'}\right) \| \ \hat{F}_{s}^{} \label{Eq4}
	\end{flalign}
	where $\|$ represents the concatenation of two feature maps along the depth axis. Note that the first convolutional layer of the $conv_{c}$ needs to reduce the feature map depth to be consistent with the other pathway feature map depth, when cascading multiple modules. \\ %making this adjustment 
	
	\noindent\textbf{Semantic Supervision. }%关于语义监督这一块，最后可以考虑是否加一下那个分支的fc结构 类似于encnet
	In practice, we come up with the deep semantic supervision in the DCA module to enhance the semantic concepts of the contextual output as connecting multiple DCA modules. To fit the DCA module to establish the long-range context fusion, we assign the semantic supervision to the pathway of concatenation, which is beneficial for improving global semantic similarity in the context information. Fig. \ref{Fig2}(b) presents the detailed structure of our Semantic Supervision block. %We present the detailed structure of our Semantic Supervision block in
	We first refine the output of the contextual pathway through a 1 $\times$ 1 $conv$ to decrease the dimension of the feature map.
	After global average pooling, we build an additional fully connected layer to implement individual predictions for the object categories in the scene 
	and learn with the binary cross-entropy loss for each category.
	%with a sigmoid activation function  The contextual pathway 
	Some similar supervision techniques \cite{Zhao_2017_CVPR}\cite{PSANet_2018}\cite{Zhang_2018_ECCV} are generally utilized with related deep networks \cite{lee2015deeply}\cite{Szegedy_2015_CVPR} to optimize the learning process.
	%such supervision techniques  
	Therefore, the semantic supervision we applied in the DCA module can not only enhance the understanding of the class-level contextual features, but also benefit the training process. The experiments in Section \ref{Exper} shows how this method improve the performance of our segmentation networks.
	%exert a good effect on training. 
	
	%On the one hand, combined with contextual information, multi-scale processing provides detailed cues and the structure's features, which can facilitate ambiguous classification due to only focus on the local context.
	%Meanwhile, taking into account that the attention module captures the contextual information by processing the input features with context scale pooling, we can flexibly utilize different scales in the DCA module.
	%这里可以替换掉这个词 , 或者这里说更好的场景理解
	%{\color{red}On the other hand, to capture the long-range contextual relationships for better improving dense feature representation, we further take a progressive approach by cascading multiple DCA modules.}
	%Specifically, inspired by the network structure of multi-scale \cite{2017arXiv170605587C}, we adjust the context pooling scale of the DCA module and vary the number of attention modules.
	\subsection{Long-range DCA structure}
	\label{sec:3.3}%这里的引用可以加上 progressively
\begin{figure}[!ht]
		\centering
		\includegraphics[width=0.9\linewidth]{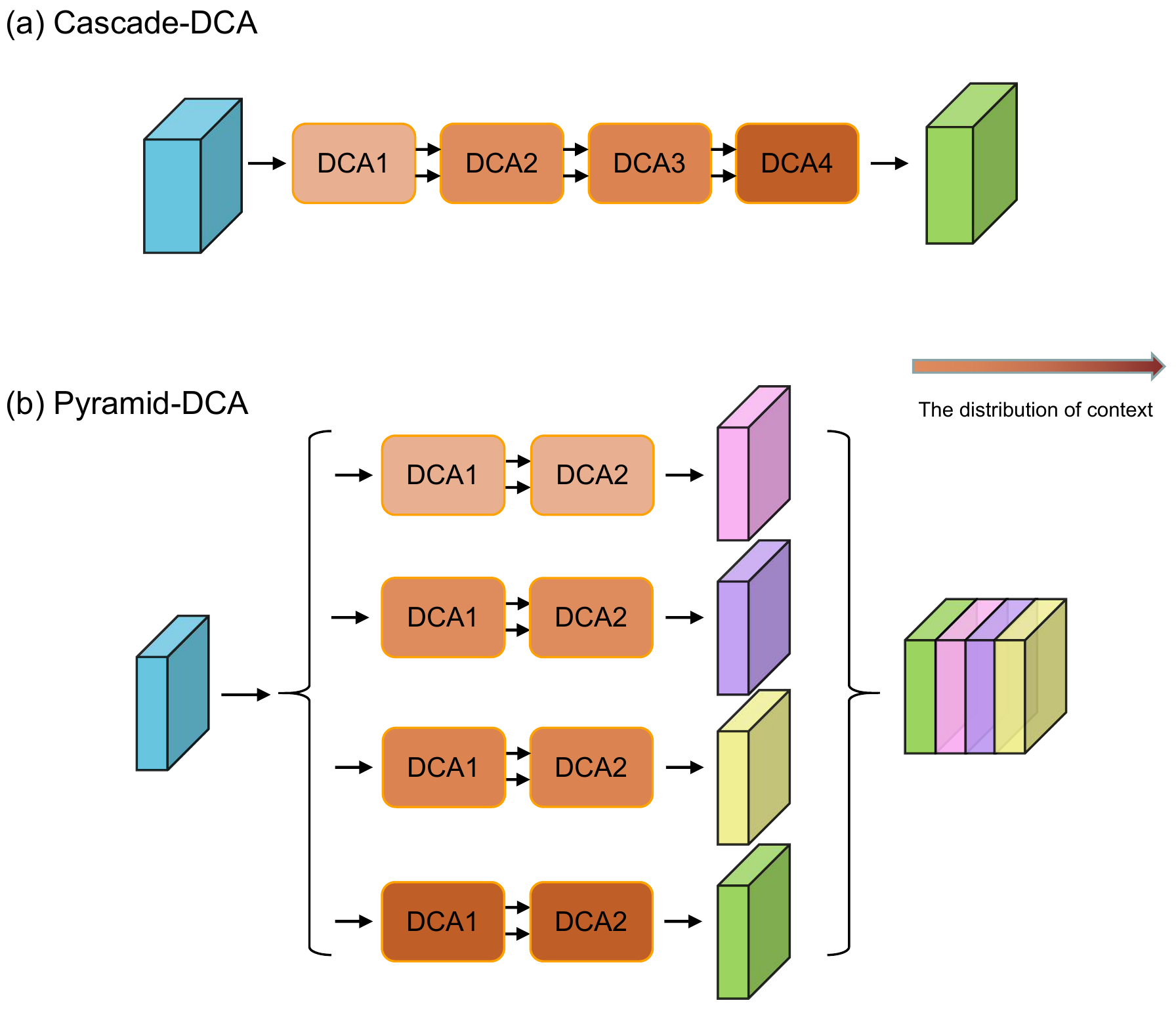}
		% figure caption is below the figure  %  对于金字塔 那个颜色我们要改成从上到下从深到浅色
		\caption{Illustration of two long-range DCA structures. %The structures of Cascade-DCA and Pyramid-DCA are illustrated in (a) and (b).
			(a) The Cascade-DCA structure: We employ four modules cascaded in sequence to construct this structure. 
			(b) The Pyramid-DCA structure: The structure consists of four parallel branches independently, and we concatenate the four output feature maps with a final feature map.
			In the cascade structure, we take the input feature map as the two inputs of the first DCA module and the spatial pathway of the last module as the final output in each cascade line.
			%The T module represents scale transformation, which produces global-scale features and local-scale features, respectively. 
			Here, we apply gradient colors to distinguish the context pooling at different scales used in the DCA module.
			%每一条级联线上的第一个模块的输入
			%We distinguish the two streams by the corresponding colors to represent the different scale features they carry. 
			(Best viewed in color)}
		%The global-context stream (blue) stays. The localization stream (red) stays. }
		\label{Fig3}       % Give a unique label
	\end{figure}

	We have introduced a separate dense context-aware module in detail, whose contextual pathway only works on one scale context. 
	Next, we present the long-range DCA structures, which comprise a sequence of the DCA modules and refine the feature maps progressively.
	On the one hand, to capture the long-range contextual relationships for better improving dense feature representation, we implement a progressive strategy \cite{Zhu_2019_CVPR} by the DCA modules cascaded in sequence.
	Correspondingly, proper network depth can be obtained by varying the number of modules.
	%we vary the number of modules for different network structures. % control the number of modules according to different network forms
	On the other hand, inspired by the multi-scale network structure \cite{Zhao_2017_CVPR}\cite{Chen_2017_CVPR}, we also enhance the capabilities of network structure for multi-scale processing, which can facilitate ambiguous classification due to only focusing on the local context. %Specifically
	By adjusting the pixel size of scale context pooling for each DCA module (see Fig. \ref{Fig2}(b)), hierarchical pyramid scales are composed when using several modules. 
	We describe the two structures in the following, and the architectures of Cascade-DCA and Pyramid-DCA are shown in Fig. \ref{Fig3}. \\
	%Given the intermediate features $X \in \mathbb{R}^{C \times H \times W}$, where $N$ is the number of the channels, we propose two network structures to complete the semantic segmentation task.
	
	%这个是串联，在这个里边说 用哪一个流的输出作为 这里可以说一下, 这里就直接说用了四个 下边金字塔说为了保证计算
	\noindent\textbf{Cascade-DCA. }
	The core of the cascade-DCA structure, consisting of four cascaded DCA modules, is to adopt a progressive strategy to cope with the understanding of segmentation scene adaptively.
	Specifically, to achieve the optimization of contextual attention from semantic level to detailed cues, we gradually increase the size $r_{l}$ of context pooling in each module, using four pyramid scales: $1 \times 1$, $4 \times 4$, $8 \times 8$ and $16 \times 16$. 
	%We feed the intermediate features to both pathways of the first module, and then through these cascaded modules, the spatial pathway of the last module is taken as the final result. %Starting from the first module, we feed the intermediate feature Meanwhile, the cascade-DCA structure 
	%Besides, to ensure the long-range contextual relationship, 
	Taking the progressive manner allows each DCA module to improve the local feature representations under the different contextual awareness, which further ensures the long-range contextual relationship.
	We feed the intermediate features to both pathways of the first module.
	Through the continuous optimization of these cascaded modules, the features of the spatial pathway have enhanced the similarity of the semantic categories while maintaining rich spatial details, which helps to boost feature discriminability for classification.
	Therefore, the spatial pathway of the last module is taken as the final result.
	More details of this Cascade-DCA structure is illustrated in Fig. \ref{Fig3}. \\
	%{\color{red}Starting from $F_{0}$, we gradully update the two pathways of the DCA module, A and B, and finally we use the spatial output of the last attention module as the results of the cascaded network structure.}   
	%Meanwhile, the size of context pooling in these attention modules also gradually increase, using four scales: 1,2,3,4, to enhance the impact of the context on local information.
	%In order to ensure the performance of network structure with low computation complexity, we 
	
	\noindent\textbf{Pyramid-DCA. }
	We employ four parallel branches, each of which consists of several DCA modules.
	%Same as the Cascade-DCA structure, we use the cascaded modules in each branch, but differently, all modules in each branch use the identical size of context pooling.
	Same as the Cascade-DCA structure, we use the cascaded modules in each branch, with all modules in each branch using the identical size of context pooling. Inspired by PSPNet \cite{Zhao_2017_CVPR}, we adopt four pyramid scales: $1 \times 1$, $2 \times 2$, $3 \times 3$, and $6 \times 6$, for the sizes $r_{l}$ of context pooling of the four branches.
	%The context pooling scale of DCA module input for four branch using four pyramid scales: $1 \times 1$, $1 \times 1$, $1 \times 1$, $1 \times 1$, respectively, which is , and 
	%%这句带颜色的话考虑换成ocnet那句feeding
	%{\color{red}considering the depth of the network, we use two DCA module in each branch,} then we apply the deep semantic supervision for the last module of each branch.
	%Meanwhile, considering the excellent performance of network structure with low computation complexity, we use two cascaded DCA modules in each branch and only apply the deep semantic supervision for the last module of each branch.
	Meanwhile, considering the excellent performance of network structure with low computation complexity, we utilize two cascaded DCA modules in each branch to take full advantages of long-range contextual information.
	Finally, the spatial pathway outputs of the last module of the four branches are concatenated together to achieve cross-branch feature fusion. More details are illustrated in Fig. \ref{Fig3}.
	%enhance the discriminative and comprehensive capability of features. 
	%And the method, called Pyramid-DCA, refers to the parallel pyramid scales. 
	%Finally, we concatenate the multiple pyramid object context representations with the input feature map.
	\subsection{Network Architecture}
	\label{sec:3.4}
	%这里首先基于long-range结构with xxx, 这里可以画个图吗？
	With the two long-range DCA structures (Cascade-DCA and Pyramid-DCA), we propose the end-to-end network for semantic segmentation.
	Regardless of the long-range structure, our network is mainly composed of two separate parts, the backbone network and the DCA structure.
	%{\color{red}It is noted that our DCA structures are scalable and can be inserted in the existing FCN pipeline effectively. } \\  %strengthen feature representations. 
	It is worth noting that our module can be embedded into the existing FCN pipeline for exploiting different network variants. \\
	
	\noindent\textbf{Baseline Network. }
	As for the baseline network, we use the ResNet-101 pre-trained on the ImageNet dataset \cite{article2012}. 
	And following \cite{chen_2018_deeplabv2}, we make some modifications: remove the classification layer and last pooling layer, and replace the convolutions within the last two modules by dilated convolutions with dilation rates being 2 and 4, respectively. Then the output feature map size is $1/8$ of the input image.\\
	
	\noindent\textbf{Long-range DCA structure. }
	%{\color{red}With the backbone network, we construct the Cascade-DCA network and the Pyramid-DCA network.} %({\color{red}corresponding to four pooling scales:})
	We construct the entire network structures based on Cascade-DCA and Pyramid-DCA, respectively.
	
	The detailed architecture of Cascade-DCA network is given as follows. %For the Cascade-DCA network, 
	We directly feed the output feature map of the backbone network into the Cascade-DCA structure which consists of four DCA modules connected in sequence and the dimension of the input feature map is reduced from 2048 to 512 in the first module of the cascaded network.
	%这里要不要写我们concat 最后两个模块的
	After going through the cascaded network, we further employ a $1 \times 1$ convolution on the output feature map with 1024 channels to obtain the final result.
	
	For the Pyramid-DCA network, we first apply a $3 \times 3$ convolution layers (with batch normalization and ReLU layers) to reduce the output dimension of backbone from 2048 to 512 in advance, then we feed the features of dimension reduction into the Pyramid-DCA structure which contains four different branches, and we concatenate the four different spatial pathway outputs from the four parallel cascaded branches. %of the last module by
	Each of the four branches of output feature maps has 512 channels. We employ a $1 \times 1$ convolution on the concatenated feature map with 2048 channels to generate the final feature map with 512 channels. \\
	
	\noindent\textbf{Loss Function. }
	In addition to the main supervision applied to the final output of our network, we employ the auxiliary supervision, and the deep semantic supervision in the DCA modules.  %这里可能要说一下用的都是 cross entropy
	For explicit feature refinement, we use extra deep supervision to refine the performance of the FCN backbone and make the network easier to optimize following PSANet \cite{PSANet_2018}. 
	For introducing the semantic information into related features, we apply the same deep semantic supervision for the last module in the Cascade-DCA and the last module of each branch in the Pyramid-DCA.
	%For introducing the semantic information into related features, we use the deep semantic supervision in the last DCA modules.
	The class-balanced cross entropy is employed for main segmentation loss, auxiliary loss, and semantic supervision losses.  
	Finally, we use three parameters $\lambda_{m}, \lambda_{a} \text { and } \lambda_{s}$ to balance the main segmentation loss $l_{m}$, the auxiliary loss $l_{a}$ and all semantic supervision losses $l_{s}$ as shown in Eq. \ref{Eq5}.
	\begin{flalign}
	L=\lambda_{m} \cdot l_{m}+\lambda_{a} \cdot l_{a}+\lambda_{s} \cdot l_{s} \label{Eq5}
	\end{flalign}
	
	\section{Experiments}
	\label{Exper}
	%{\color{red}Our proposed approach is effective on the semantic segmentation challenges.}   
	To evaluate the proposed approach, we carry out comprehensive experiments on three challenging datasets: object segmentation dataset PASCAL VOC 2012 \cite{Everingham2015}, Cityscapes dataset \cite{Cordts_2016_CVPR}, and ADE20K \cite{Zhou_2017_CVPR}.  
	In the following, we first introduce the implementation details related to training strategies on different datasets and hyper-parameters, then we report experimental results and ablation study on corresponding datasets.
	Finally, we present the progressive aggregation processing by the visualization of the learned masks generated by the DCA modules.
	%Finally, we discuss some results related to the DCA module.
	%{\color{red}It is worth noting that due to the restrictions of GPU resources, our experiments adjust some operations, which limits the performance to some extent.}
	
	\subsection{Implementation Details}
	We implement our experiments based on Pytorch.
	Following prior works \cite{chen_2018_deeplabv2}, we adopt a \textit{poly} learning rate policy where the initial learning rate is multiplied by $\left(1-\frac{iter}{max_{i} ter}\right)^{power}$. The initial learning rate is set to 0.01 for Cityscapes dataset and 0.001 for others, and the power is set to 0.9. 
	%这里batchsize可以根据不同数据集不一样
	We train our model with mini-batch stochastic gradient descent (SGD) \cite{article2012} and set the batch size to 8 for Cityscapes and 16 for others, the momentum to 0.9, and the weight decay to 0.0001, respectively.
	The performance of the model can be improved by increasing the number of iterations, which is set to 30K for PASCAL VOC, 90K for Cityscapes and 150K for ADE20K.
	For data augmentation, we employ the random mirror, and random resize between 0.5 and 2.0 for all datasets and additionally add new random rotation between -10 and 10 degrees and random Gaussian blur for PASCAL VOC 2012 and ADE20K datasets.
	%Then we also add extra random rotation between -10 and 10 degrees and random Gaussian blur for PASCAL VOC 2012 and ADE20K datasets. 
	We notice that data augmentation does help improving performance and avoiding overfitting.
	In the experiments, the loss weights $\lambda_{m}, \lambda_{a} \text { and } \lambda_{s}$ in Eq. \ref{Eq5} are set to 1.0, 0.2 and 0.05 respectively.
	%As for the auxiliary loss, we finally set the weight $\lambda$ to 0.4 after a series of comparison experiments.
	%In the experiment, we set the weight $\lambda$ to 0.4 for the auxiliary loss.
	
	\subsection{PASCAL VOC 2012}
	\noindent\textbf{Dataset and Evaluation Metrics. }
	We perform a series of experiments on the PASCAL VOC 2012 segmentation dataset, which is for object-centric segmentation and contains 20 object classes and one background. 
	Following prior works \cite{Chen_2015_ICLR}\cite{papandreou2015weakly}, we use the augmented annotations from \cite{Hariharan2011} resulting 10,582, 1,449 and 1,456 images for training, validation and testing. For evaluation metrics, the mean of class-wise intersection over union (Mean IoU) is adopted.
	
	\subsubsection{Ablation Study}
	We first use the atrous ResNet-101 as the backbone network, and the final segmentation results are obtained by directly upsampling the output.
	For starters, we evaluate the performance of the baseline network and conduct experiments based on the backbone (ResNet-101).
	It should be noted that all our experiments use the auxiliary supervision to optimize the learning process. \\ %and deep semantic supervision
	
	\noindent\textbf{Dense Context-Aware module. }
	We perform two comparison experiments to evaluate the effectiveness of our network structure with the DCA module.
	%To evaluate the effectiveness of our network structure with the DCA module, we carry out two comparison experiments.
	One is to explore the advantage of the DCA module in the Cascade-DCA structure by replacing it with the vanilla residual module \cite{He_2016_CVPR}\cite{johnson2016perceptual} that results in a cascaded residual structure named ResNet-101 + CRS. Specifically, we use ResNet-101 and multiple residual modules, and the number of these residual modules is the same as in the Cascade-DCA structure.
	And the other is to exhibit the advantage of our Pyramid-DCA structure with the DCA modules. Similarly, for comparison, we use ResNet-101 + PPM to represent the PSPNet that applies pyramid pooling module on feature maps of multiple scales. %这里要说一个尺度保持一致可以放在表格注释下边
	The related experimental results with different settings are reported in Table \ref{tab1}, where the single scale testing is adopted in all the results.
	Especially, the performance of these methods has indicated the mean through several times to ensure that our results are reliable.
	
	As shown in Table \ref{tab1}, our approaches with the DCA modules outperform the baseline network remarkably.
	Compared with the ResNet-101 + CRS, employing the cascaded modules in the Cascade-DCA structure yields a result of 77.45\% in Mean IoU, which increases by almost 3\%.
	Meanwhile, employing the Pyramid-DCA structure with the DCA modules exceeds the individual pyramid scale structure by 1.9\%.
	Notably, our Pyramid-DCA network improves the segmentation performance over the Cascade-DCA network by 0.6\% on the validation set, which shows that the former is a slightly better choice in terms of capturing long-range contextual information. The complexity of the Cascade-DCA structure is relatively small, however, and only requires about 1/2 of the parameters of the Pyramid-DCA structure. Considering the balance between performance and complexity, the Cascade-DCA network is also a solution worth exploring.\\
	\begin{table}[ht]
		\centering
		\caption{Detailed performance comparison of our proposed networks with different approaches on the validation set of PASCAL VOC 2012.
			Results are reported with the same settings. %All the set based on the ResNet-101. %后面可以跟 denote or 冒号
		} %添加标题 设置标签
		\label{tab1}
		\begin{tabular}{c|cc|c}
			\toprule[1.5pt]
			Method & Mean IoU(\%)\\
			\midrule\midrule
			ResNet-101 Baseline  & 73.64 \\
			\midrule
			ResNet-101 + CRS &74.56\\
			ResNet-101 + PPM  &76.14\\ %75.84
			\midrule
			ResNet-101 + Cascade-DCA &77.45\\
			ResNet-101 + Pyramid-DCA  &78.06\\
			\bottomrule[1.5pt]
		\end{tabular}
	\end{table}
	
	\noindent\textbf{Ablation Study for Improvement Strategies. }
	Following \cite{Chen_2017_CVPR}, we adopt some strategies to improve the performance of the network further. These improvement strategies are: DA (Data augmentation with random scaling), MS (We average the segmentation probability maps from 7 image scales \{0.5 0.75 1 1.25 1.5 1.75 2\} for inference), and SS (the semantic supervision loss in the corresponding DCA modules). %{\color{red}Multi-Grid (we apply a hierarchy of grids of different sizes (4,8,16) in the last ResNet module)} %这是实在不行就改成多网格
	
	We conduct experiments on the basis of the ResNet-101 + Cascade-DCA structure, and the results are reported in Table \ref{tab2}.
	From the experimental results, we notice that using data augmentation with scaling improves the performance by almost 1.0\% because of the enriching scale diversity of training data.
	Also, using the semantic supervision, our result can further exceed it by 1.5\% and reach 79.95\%, which shows that network benefits from the deep semantic supervisions' capability of enhancing the context fusion.
	Finally, by applying the multi-scale testing, segmentation result fusion further improves the performance to 80.84\%, which outperforms the original method by 3.4\%. 
	
	\begin{table}[ht]
		\centering
		\caption{Performance comparison between different improvement strategies on PASCAL VOC 2012 val set. 
			DCANet-101 represents DCANet with the backbone ResNet-101, 'DA' represents data augmentation with random scaling. 
			'SS' represents employing the semantic supervision, 'MS' represents multi-scale inputs during inference.} %添加标题 设置标签
		\label{tab2}
		\begin{tabular}{c|ccc|c}
			\toprule[1.5pt]
			Method& DA & SS & MS &Mean IoU(\%)\\
			\midrule\midrule
			Baseline(Res101) & & & &73.64\\
			DCANet-101 & & & &77.45\\
			DCANet-101 &$\surd$ & & &78.68\\
			DCANet-101 &$\surd$ &$\surd$ & &79.95\\
			DCANet-101 &$\surd$ &$\surd$ &$\surd$ &80.84\\
			\bottomrule[1.5pt]
		\end{tabular}
	\end{table} 
	
	\subsubsection{Method Comparison}
	%这里完全可以两个方法都用上, t and pra
	We show the comparison between our method and some previous methods in Table \ref{tab3}.   
	With pre-training on the ImageNet dataset, our method based on ResNet-101 achieves the superiority over these previous state-of-the-art methods\footnote{The result link to the VOC evaluation server: \\ http://host.robots.ox.ac.uk:8080/anonymous/B3XPSK.html}.
	In particular, our model is better than some methods, such as DeepLabv2-CRF \cite{chen_2018_deeplabv2} and GCN \cite{Peng_2017_CVPR}, which use powerful pretrained models on the MS-COCO dataset.
	Comparing to state-of-the-art approaches of DANet \cite{Fu_2019_CVPR} and SANet \cite{Zhong_2020_CVPR}, our method improves the performance to 84.41\%.
	%has less computation complexity
	Furthermore, we believe that our proposed approach (e.g., the DCA module and the cascaded structure) could be useful for many vision tasks.
	%\noindent\textbf{Visual Improvements. }
	%\begin{table}[ht]
	%	\centering
	%	\caption{Methods comparison with results reported on PASCAL VOC 2012 testing dataset.
	%		Methods pre-trained on MS-COCO are marked with $^{'}$.}
	%	%Methods trained on our devices are marked with $^{'}$. } %添加标题 设置标签
	%	\label{tab3}
	%	\begin{tabular}{l|cc|c}
	%		\toprule[1.5pt]
	%		Method&Mean IoU(\%)\\
	%		\midrule\midrule %\midrule[1.5pt]
	%		CRF-RNN$^{'}$ \cite{Zheng_2015_ICCV} &74.7\\
	%		DPN$^{'}$ \cite{Liu_2015_ICCV}&77.5\\
	%		Piecewise$^{'}$ \cite{Lin_2016_CVPR}&78.0\\
	%		DeepLabv2-CRF$^{'}$ \cite{chen_2018_deeplabv2}& 79.7 \\
	%		SegModel$^{'}$ \cite{Shen_2017_CVPR}&81.8 \\
	%		DUC\_HDC$^{'}$ \cite{Zhao_2017_CVPR} &83.1\\
	%		GCN$^{'}$ \cite{Peng_2017_CVPR}&83.6 \\
	%		PSPNet \cite{Zhao_2017_CVPR} &82.6\\
	%		DFN \cite{Yu_2018_CVPR} &82.7\\
	%		EncNet \cite{Zhang_2018_CVPR} &82.9\\
	%		DANet \cite{Fu_2019_CVPR} &82.6\\
	%		\midrule
	%		DCANet &84.0\\
	%		\bottomrule[1.5pt]
	%	\end{tabular}
	%\end{table} 
	\begin{table}[ht]
		\centering
		\caption{Methods comparison with results reported on PASCAL VOC 2012 testing dataset.
			Methods pre-trained on MS-COCO are marked with $^{'}$.}
		%Methods trained on our devices are marked with $^{'}$. } %添加标题 设置标签
		\label{tab3}
		\begin{tabular}{l|c|c}
			\toprule[1.5pt]
			\textbf{Method}&\textbf{Backbone}&\textbf{Mean IoU(\%)}\\
			\midrule\midrule %\midrule[1.5pt]
			FCN\cite{Shelhamer2017} &&62.2\\
			CRF-RNN \cite{Zheng_2015_ICCV} &&72.0\\
			DPN \cite{Liu_2015_ICCV}&&74.1\\
			%		Piecewise \cite{Lin_2016_CVPR}&&75.3\\
			DeepLabv2-CRF$^{'}$ \cite{chen_2018_deeplabv2} &ResNet-101 & 79.7\\
			GCN$^{'}$ \cite{Peng_2017_CVPR}&ResNet-101&83.6 \\
			ResNet38 \cite{wu2019wider}&WideResNet-38&82.5\\
			RefineNet \cite{Lin_2017_CVPR}&ResNet152&83.4\\
			PSPNet \cite{Zhao_2017_CVPR} &ResNet-101&82.6\\
			EncNet \cite{Zhang_2018_CVPR} &ResNet-101&82.9\\
			DANet \cite{Fu_2019_CVPR} &ResNet-101&82.6\\
			%		CFNet \cite{Zhang_2019_CVPR} &ResNet-101&84.2\\
			APCNet \cite{He_2019_CVPR} &ResNet-101&84.2\\
			SANet \cite{Zhong_2020_CVPR} &ResNet-101&83.2\\
			%		DFN \cite{Yu_2018_CVPR} &ResNet-101&82.7\\
			\midrule
			DCANet &ResNet-101&84.4\\
			\bottomrule[1.5pt]
		\end{tabular}
	\end{table} 
	\subsubsection{Visual Improvements}       % 也可以具体到物体说的细一点  类比PSA  
\begin{figure}[!ht]
		\centering
		\includegraphics[width=1.0\linewidth]{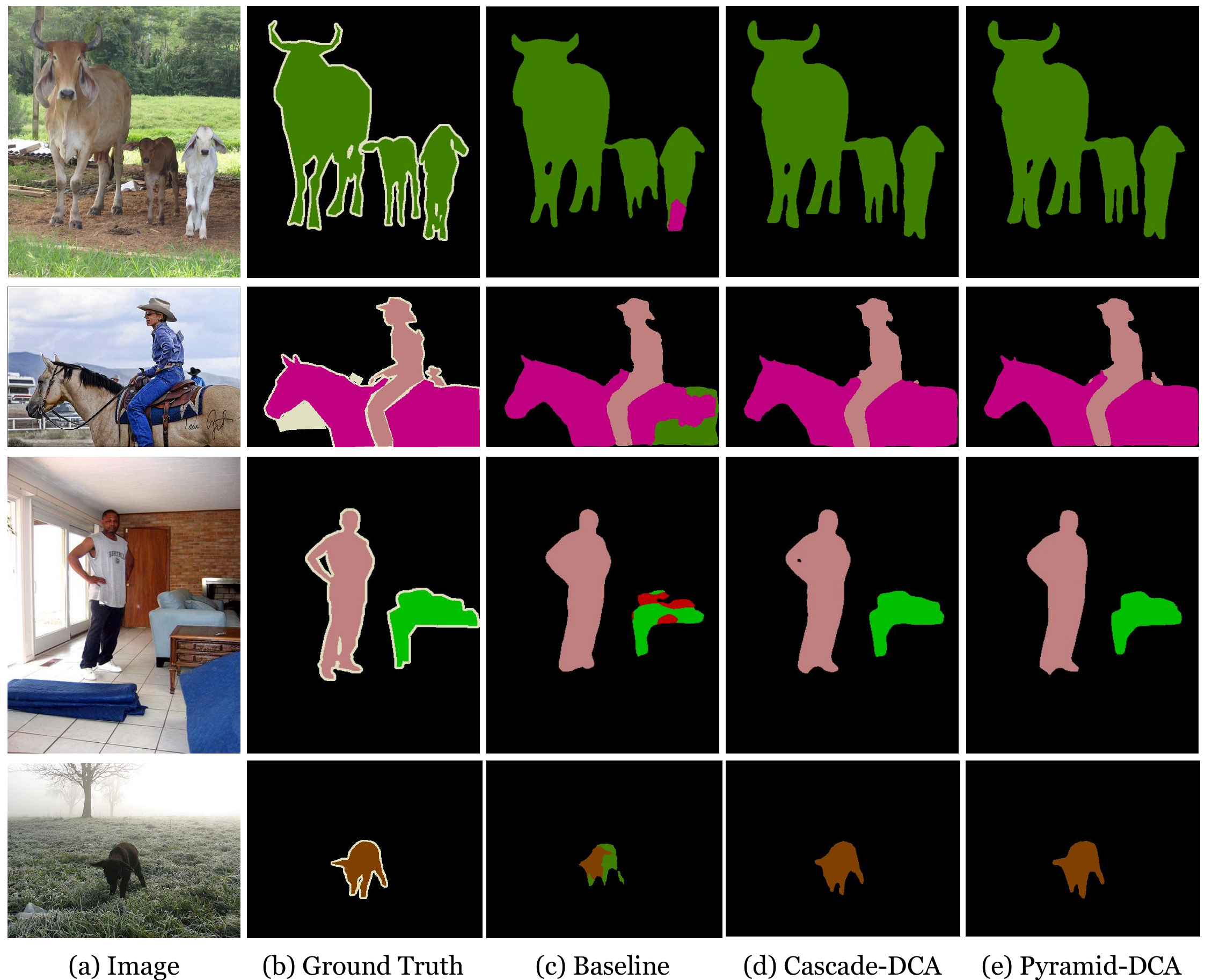}
		\caption{Visual improvement on validation set of PASCAL VOC 2012. Our proposed methods get
			more accurate and detailed segmentation results. 
			'Cascade-DCA' denotes our DCANet with the Cascade-DCA structure and 'Pyramid-DCA' represents DCANet with the Pyramid-DCA structure, which further enhances the prediction.}
		\label{Fig4}       % Give a unique label
	\end{figure}

	To further demonstrate the effectiveness of our method, we show the visual comparison of the segmentation results in Fig. \ref{Fig4}.
	Consistently, our methods improve the segmentation quality, where more accurate and detailed structures are obtained compared to the baseline.
	Some misclassified categories are now correctly classified, such as the \textit{horse} in the second row and the \textit{sofa} in the third row.
	Meanwhile, we find out that some details and object boundaries are clearer from the results, such as the \textit{cow} in the lower right of the first row.
	%Meanwhile, we can find out from the results that local details, such as the \textit{cow} in the lower right of the first row.
	Besides, in terms of integrity and boundary details, slightly better segmentation maps than Cascade-DCA are produced by the Pyramid-DCA.
	
	\subsection{Cityscapes}
	\noindent\textbf{Dataset. }
	Cityscapes dataset \cite{Cordts_2016_CVPR} is collected for semantic segmentation on urban street scenes.
	It contains 5,000 finely annotated images captured from 50 cities in different seasons. And these images are divided into 2,975, 500, and 1,525 images for training, validation and testing. 
	It has 30 annotated common classes of road, person, car, etc. and defines 19 classes for semantic segmentation evaluation.
	Besides, another 20,000 coarsely annotated images are also provided.
	
	%可以模仿下DANet PASCAL VOC 2012
	We carry out experiments on the Cityscapes dataset to evaluate the effectiveness of our method.
	We first show the improvement brought by our DCANet based on the FCN backbone with different layers (ResNet-50 or ResNet-101). Note that we also adopt 
	different DCA structures in Table \ref{tab4}.
	The baseline (ResNet-50) yields Mean IoU 72.33\%. Our DCA structure improves performance significantly, where DCANet-50 for better results exceeds the baseline by 6.6\%.
	When we adopt a deeper network ResNet-101, the model achieves Mean IoU 80.13\%. 
	%some improvement strategies () to improve the performance of the DCANet  % 这里可以在多说一些关于提升 %xxx
	To further illustrate the performance of our method on the Cityscapes dataset, we show the comparison with some previous methods.
	The evaluation results of Cityscapes val set are shown in Table. \ref{tab5}, and our method achieves the best performance under both settings.
	\begin{table}[ht]
		\centering
		\caption{Performance comparison between different strategies on Cityscapes val set. Results are reported with
			models for single-scale testing and. } %添加标题 设置标签
		\label{tab4}
		\begin{tabular}{c|ccc|c}
			\toprule[1.5pt]
			Method& BaseNet & Cascade-DCA & Pyramid-DCA & Mean IoU(\%)\\
			\midrule\midrule
			Baseline &ResNet-50 & & &72.33 \\
			DCANet &ResNet-50 &$\surd$ & &78.05 \\
			DCANet &ResNet-50 & &$\surd$ &78.92 \\
			\midrule
			Baseline &ResNet-101 & & &74.72 \\
			DCANet &ResNet-101&$\surd$ & &79.55 \\
			DCANet &ResNet-101& &$\surd$ &80.13 \\
			\bottomrule[1.5pt]
		\end{tabular}
	\end{table} 
	
	\begin{table}[ht]
		\centering
		\caption{Methods comparison with results reported on Cityscapes set. 
			We adopt the same improvement strategies as in PASCAL VOC 2012 to improve performance.
			%Methods trained on our devices are marked with $^{'}$.
			%Methods trained using both fine and coarse data are marked with $^{'}$. 
		} %添加标题 设置标签 'DA' refers to data augmentation we performed.
		\label{tab5}
		\begin{tabular}{l|cc|c}
			\toprule[1.5pt]
			Method&Mean IoU(\%)\\
			\midrule\midrule
			FCN \cite{Shelhamer2017} &65.3\\
			DeepLabv2-CRF \cite{chen_2018_deeplabv2}& 70.4 \\
			RefineNet \cite{Lin_2017_CVPR}&73.6\\
			%		GCN$^{}$\cite{Peng_2017_CVPR}&76.2\\
			DUC\_HDC \cite{wang2018understanding}&76.9\\
			PSPNet$^{}$ \cite{Zhao_2017_CVPR} &77.7\\
			%		SegModel$^{}$ \cite{Shen_2017_CVPR}&77.8\\
			%		SegModel$^{}$ \cite{Shen_2017_CVPR}&77.8\\
			%		DFN$^{}$ \cite{Yu_2018_CVPR} &78.6\\
			PSANet$^{}$ \cite{PSANet_2018} &79.3\\
			DenseASPP \cite{Yang_2018_CVPR} & 79.8\\
			CCNet \cite{Huang_2019_ICCV} &81.3\\
			DANet \cite{Fu_2019_CVPR} &81.5\\
			\midrule
			DCANet &81.8\\
			\bottomrule[1.5pt]
		\end{tabular}
	\end{table}
	
\begin{figure}[htbp]
		\centering
		\includegraphics[width=1.0\linewidth]{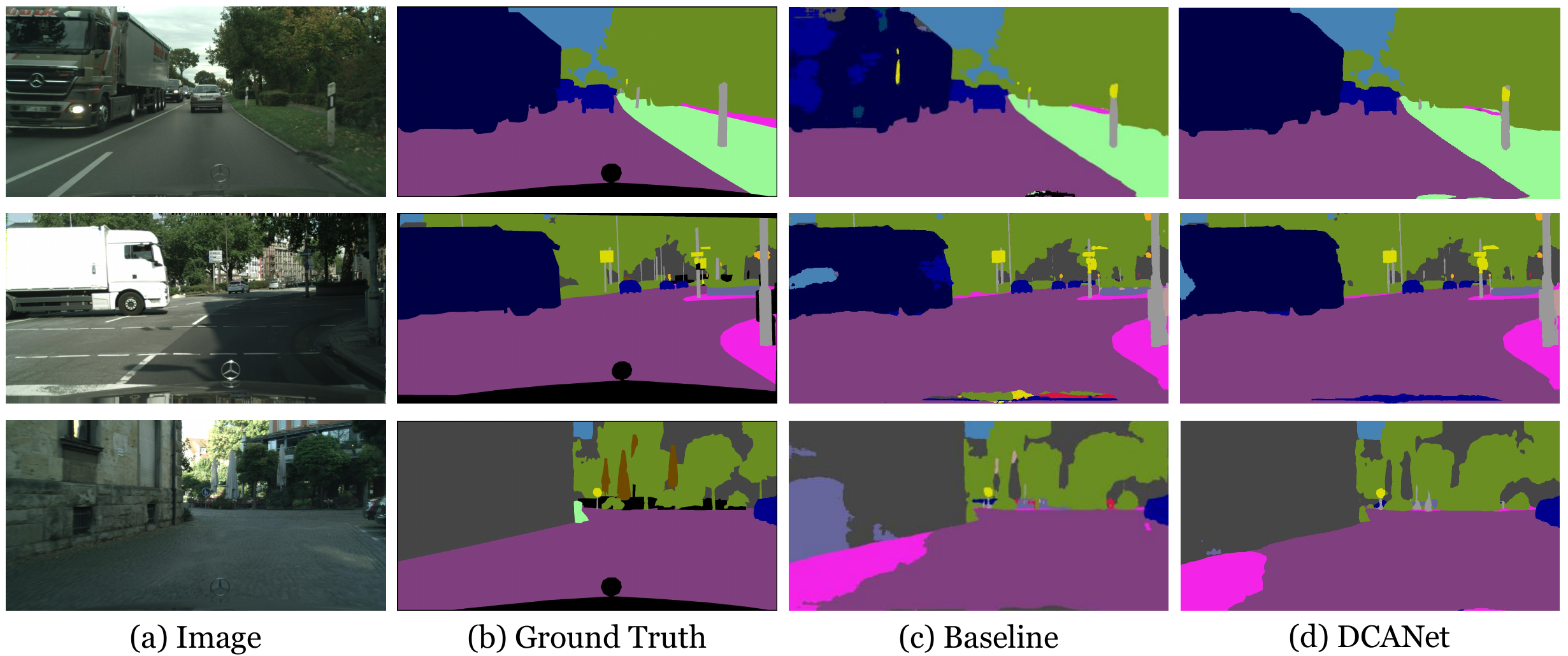}
		\caption{Examples of DCANet results on Cityscapes dataset. } %Our DCANet produces more accurate and detailed results.
		\label{Fig6}       % Give a unique label
	\end{figure}

	%To present our experimental results, several examples with visual improvement are clear, as shown in Fig. \ref{Fig6}.
	We also provide the qualitative comparisons between DCANet and baseline network on several examples to present the visual improvement, as shown in Fig. \ref{Fig6}.
	Similarly, better prediction is yielded with the DCA structure incorporated.
	For the parsing of complex scenes such as the traffic street, our network gets better performance than the baseline when dealing with objects of the various scales.

	\subsection{ADE20K}
	\noindent\textbf{Dataset. }
	The ADE20K dataset \cite{Zhou_2017_CVPR} is a scene parsing dataset, which contains 150 classes and diverse complex scenes up to 1,038 image-level categories. 
	The challenging dataset is divided into 20K/2K/3K for training, validation and testing, respectively.
	Note that both objects and stuffs need to be parsed in this dataset.
	For evaluation metrics, both \textit{pixel-wise accuracy} (Pixel Acc.) and \textit{mean of class-wise intersection over union} (Mean IoU) are used.
	
	We perform experiments to verify the generalization of our proposed network on the ADE20K dataset.
	The comparisons with some previous methods are reported in Table \ref{tab6}.
	Under the same settings, we can observe that the DCANet (ResNet101 +  Pyramid-DCA) overall achieves better results than these previous works on the validation set of ADE20K. 
	%outperforms some previous works. improves 1.6\% over the previous methods 
	Our method could capture the long-range contextual information effectively for more accurate segmentation results. 
	
	\begin{table}[ht]
		\centering
		\caption{Methods comparison with results reported on ADE20K validation dataset. } %添加标题 设置标签%Methods trained on our devices are marked with $^{'}$.
		\label{tab6}
		\begin{tabular}{l|cc}
			\toprule[1.5pt]
			Method&Mean IoU(\%) &Pixel Acc.(\%)\\
			\midrule \midrule %[1.5pt]
			FCN \cite{Shelhamer2017} &29.39 &71.32\\
			SegNet \cite{Badrinarayanan2017} &21.64 &71.00\\
			%		DilatedNet \cite{Yu_2016_ICLR} & 32.31 &73.55 \\
			CascadeNet \cite{Zhou_2017_CVPR} & 34.90 &74.52 \\
			RefineNet \cite{Lin_2017_CVPR} & 40.20 &- \\
			PSPNet \cite{Zhao_2017_CVPR} &43.29 &81.39 \\
			%		WiderNet$^{}$ \cite{chen_2018_deeplabv2}& 43.73 &81.17\\
			DSSPN \cite{Liang_2018_CVPR}& 43.68 &81.13\\
			PSANet$^{}$ \cite{PSANet_2018} &43.77 &81.51\\
			EncNet \cite{Zhang_2018_CVPR} &44.65 &81.19\\
			CCNet \cite{Fu_2019_CVPR} &45.22 &81.61\\
			APCNet \cite{He_2019_CVPR} &45.38 &-\\
			\midrule
			DCANet &45.49 & 81.65\\
			\bottomrule[1.5pt]
		\end{tabular}
	\end{table}  %		DCANet &43.05 & 80.98\\
	%We further conduct experiments on semantic part segmentation which utilizes the extra PASCAL VOC 2010 annotations by \cite{2014arXiv1406.2031C}.
	%We focus on the $person$ part for the dataset, which contains more training data and large variation in scale and human pose.
	%Specifically, the dataset contains detailed part annotations for every person, \textit{e.g.} eyes, nose.
	%We merge the annotations to be Head, Torso, Upper/Lower Arms and Upper/Lower Legs, resulting in six person part classes and one background class. And We use those images containing persons for training (1716 images) and validation (1817 images).
	%
	%We perform experiments to evaluate the DCANet (ResNet-101 + P-DCA) on the PASCAL-Person-Part dataset, and the results are reported in Table \ref{tab4}. 
	%We can observe that the OCNet improves 1.6\% points
	%From the experimental results, we can notice that using the DCA structure improves 1.6\% points over the previous state-of-the-art methods on the validation set. 
	%Especially, the human parsing task verifies that DCANet generalizes well to the part-level semantic segmentation tasks.
	%%这里可以在表格中加入一点消融试验
	%To further illustrate our experimental results, several examples are clearly shown in Fig. \ref{Fig8}. 
	%Consistently, in terms of integrity and detailed predictions, better segmentation images are yielded with our proposed network.
	
	\subsection{Visualization of attention masks}
\begin{figure*}[!ht]
		\centering
		\includegraphics[width=1.0\linewidth]{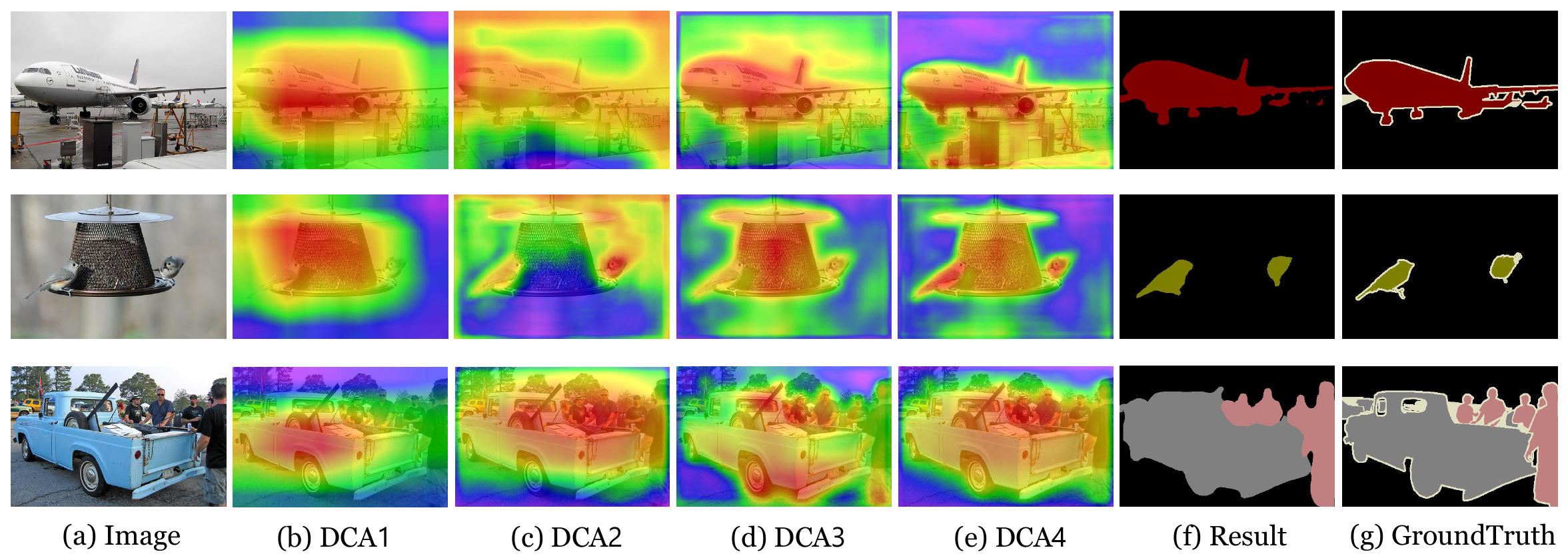}
		\caption{Visualization of learned masks by DCANet on the validation set of PASCAL VOC 2012.
			The left column is the input images from dataset, the 2, 3, 4, 5 are pixel-wise masks from corresponding modules. 
			'DCA1' denotes the attention mask of the first DCA module and 'DCA2$\sim$4' likewise.
			In addition, the corresponding result and ground-truth are provided in the last two columns. }
		\label{Fig5}       % Give a unique label
	\end{figure*}

	%In order to deeper understand how the cascaded DCA modules progressively improve the performance of the context fusion, we visualize the learned attention masks, as shown in Fig. \ref{Fig5}.
	In order to deeper understand how the progressive solution in multiple cascaded DCA modules improves the performance of the context fusion, we visualize the learned attention masks, as shown in Fig. \ref{Fig5}.
	The example images are selected from the validation set of Pascal VOC 2012.
	The progressive convergence of attention masks under the awareness of the context is clear and interpretable, which also expounds the transfer process from the global information to the details well.
	For the several masks in the front (first two columns of masks), we note that the attention tends to distinguish between foreground and background in a global view, which can help improve the capability of feature for the intra-class inconsistency problem.
	And for the last two mask columns, we find that some clear boundaries for locating the detailed clues are generated by the attention masks. In short, the visualized masks further demonstrate that collecting semantic similarity and the long-range contextual dependencies are essential for improving feature representations in semantic segmentation.
	
	%The pose attention transfer process is clear and interpretable, where the attention masks always attend to the regions needed to be adjusted in a progressive manner.
	%The initial several masks (first three columns of masks) are some sort of blending of condition pose and target pose.
	%可以说一开始的偏向于整体性，慢慢的迁移到细节上！
	%Besides, we observe that the attention module could capture semantic similarity and long-range dependencies.
	
	\section{Conclusion}
	\label{conc}
	In this work, we propose the DCA module for semantic segmentation with the objective of improving the capabilities of neural networks for the context fusion.
	Based on the DCA module, we further propose two extended structures, named Cascade-DCA structure and Pyramid-DCA structure, to progressively and adaptively capture long-range contextual information for the robustness of segmentation results.
	We demonstrate the advantages of our proposed approaches with delightful performance on three challenging benchmarks, including PASCAL VOC 2012, Cityscapes, and ADE20K.
	%Meanwhile, the ablation studies and visualization of intermediate results further enhance the interpretability of deep learning systems.
	In the future, we will concentrate on improving the computational efficiency of the dense context-aware module for semantic segmentation.
	\section{acknowledgements}
	This work was supported by the Opening Foundation of the State Key Laboratory (No. 2014KF06), and the National Science and Technology Major Project (No. 2013ZX03005013).
	
	\section*{References}
	
	\bibliographystyle{abbrv}
	\bibliography{mybibfile}
	
\end{document}